\documentclass{article}
\usepackage{spconf,amsmath,graphicx,subfigure}
\usepackage{gensymb}
\usepackage{booktabs}
\usepackage{url}
\usepackage{siunitx}
\usepackage{float}
\usepackage{multirow}
\usepackage{setspace}
\usepackage{cite}
\usepackage{amsfonts,amssymb}
\usepackage{bm}
\usepackage{hyperref}
\usepackage{hyperref}

\newcommand{\ie}{\textit{i}.\textit{e}.}

\hypersetup{hidelinks}

\title{SRF-NET: SELECTIVE RECEPTIVE FIELD NETWORK FOR ANCHOR-FREE TEMPORAL ACTION DETECTION}
\name{Ranyu Ning$^1$, Can Zhang$^1$, Yuexian Zou$^{1,2,*}$
\thanks{This paper was partially supported by the IER foundation (No. HT-JD-CXY-201904) and  Shenzhen Municipal Development and Reform Commission (Disciplinary Development Program for Data Science and Intelligent Computing). Special acknowledgements are given to Aoto-PKUSZ Joint Lab for its support. $^*$zouyx@pku.edu.cn}}
\address{$^1$ADSPLAB, School of ECE, Peking University, Shenzhen, China\\
$^2$Peng Cheng Laboratory, Shenzhen, China\\
}

\begin{document}
\ninept
\maketitle
\begin{abstract}
Temporal action detection (TAD) is a challenging task which aims to temporally localize and recognize the human action in untrimmed videos. Current mainstream one-stage TAD approaches localize and classify action proposals relying on pre-defined anchors, where the location and scale for action instances are set by designers. Obviously, such an anchor-based TAD method limits its generalization capability and will lead to performance degradation when videos contain rich action variation.
In this study, we explore to remove the requirement of pre-defined anchors for TAD methods. A novel TAD model termed as Selective Receptive Field Network (SRF-Net) is developed, in which the location offsets and classification scores at each temporal location can be directly estimated in the feature map and SRF-Net is trained in an end-to-end manner. Innovatively, a building block called Selective Receptive Field Convolution (SRFC) is dedicatedly designed which is able to adaptively adjust its receptive field size according to multiple scales of input information at each temporal location in the feature map. 
Extensive experiments are conducted on the THUMOS14 dataset, and superior results are reported comparing to state-of-the-art TAD approaches.
\end{abstract}   
%

\section{Introduction}
\label{sec:intro}
In recent years, video understanding has attracted widespread attention from academia and industry. Temporal action detection (TAD) is a fundamental yet challenging task in the video understanding area, which requires the algorithm to locate the start and end time of each action instance in an untrimmed video and also recognize the category of the action instance.

Akin to object detection approaches \cite{ren2015faster,liu2016ssd}, recent TAD approaches can be divided into two categories: two-stage approach \cite{xu2017r,chao2018rethinking,lin2018bsn,caba2016fast,gao2017turn,buch2017sst,lin2019bmn} and one-stage approach \cite{yeung2016end,lin2017single,DBLP:conf/bmvc/BuchEGFN17,long2019gaussian}. The two-stage approach first generates temporal action proposals and then classifies them in the second stage. However, the separation of proposal generation and classification may result in sub-optimal performance and the speed is usually slow due to high computational cost. On the contrary, the one-stage approach directly generates and classifies the action proposals in one stage, which is an end-to-end framework and increases inference speed. Most one-stage TAD approaches rely on pre-defined anchors and need to be set anchor-related hyper-parameters with prior knowledge of action distribution. As shown in \cite{long2019gaussian}, one-stage anchor-based detectors detect actions at predetermined temporal scales and only capture action instances whose temporal durations are well aligned to the size of receptive fields, which limits the flexibility of detecting actions with complex variations. Recently, anchor-free object detectors \cite{law2018cornernet,kong2020foveabox,tian2019fcos} have been proposed and achieve great success in the object detection task. For example, FCOS \cite{tian2019fcos} solves object detection in a per-pixel prediction fashion that predicts the distances from the center point to the four edges of the bounding box. 
These works inspire us to explore an anchor-free mechanism to improve the flexibility of the one-stage TAD method. 
Besides, as the duration of an action instance varies, a fixed receptive field is not suitable for actions of different scales, thus how to adapt the receptive field for them is also a challenging problem. The common solution is to use the feature pyramids \cite{lin2017feature} in which feature maps with different receptive fields are responsible for actions with different scales. However, this model also needs to be carefully designed and is computationally intensive. 


This work proposes a novel Selective Receptive Field Network (SRF-Net) for TAD task. SRF-Net directly predicts location offsets and classification scores at each temporal location in the feature map without pre-defined anchors, making it flexible. Meanwhile, 
inspired by the selection mechanism in SKNets \cite{li2019selective}, we delve into the selection mechanism for the receptive field to further handle scale variation.
We design a building block called Selective Receptive Field Convolution (SRFC). SRFC block is able to adaptively adjust its receptive field size according to multiple scales of input information at each temporal location in the feature map. The different receptive field at each temporal location is more effective at a specific temporal scale, which is responsible for classification and regression of the corresponding action proposal.

The main contributions of our work are summarized as follows: (1) We design a new one-stage framework for TAD task, which is anchor-free, avoiding hyper-parameters and computation related to anchors. (2) We make an exploration of the dynamic selection mechanism for the receptive field. The novel SRFC block exhibits its adaptation ability for TAD task. (3) Our proposed SRF-Net achieves superior performance than state-of-the-art TAD approaches on the standard benchmark THUMOS14 \cite{THUMOS14}.

\begin{figure}[t]
    \centering
    \includegraphics[width=0.95 \linewidth]{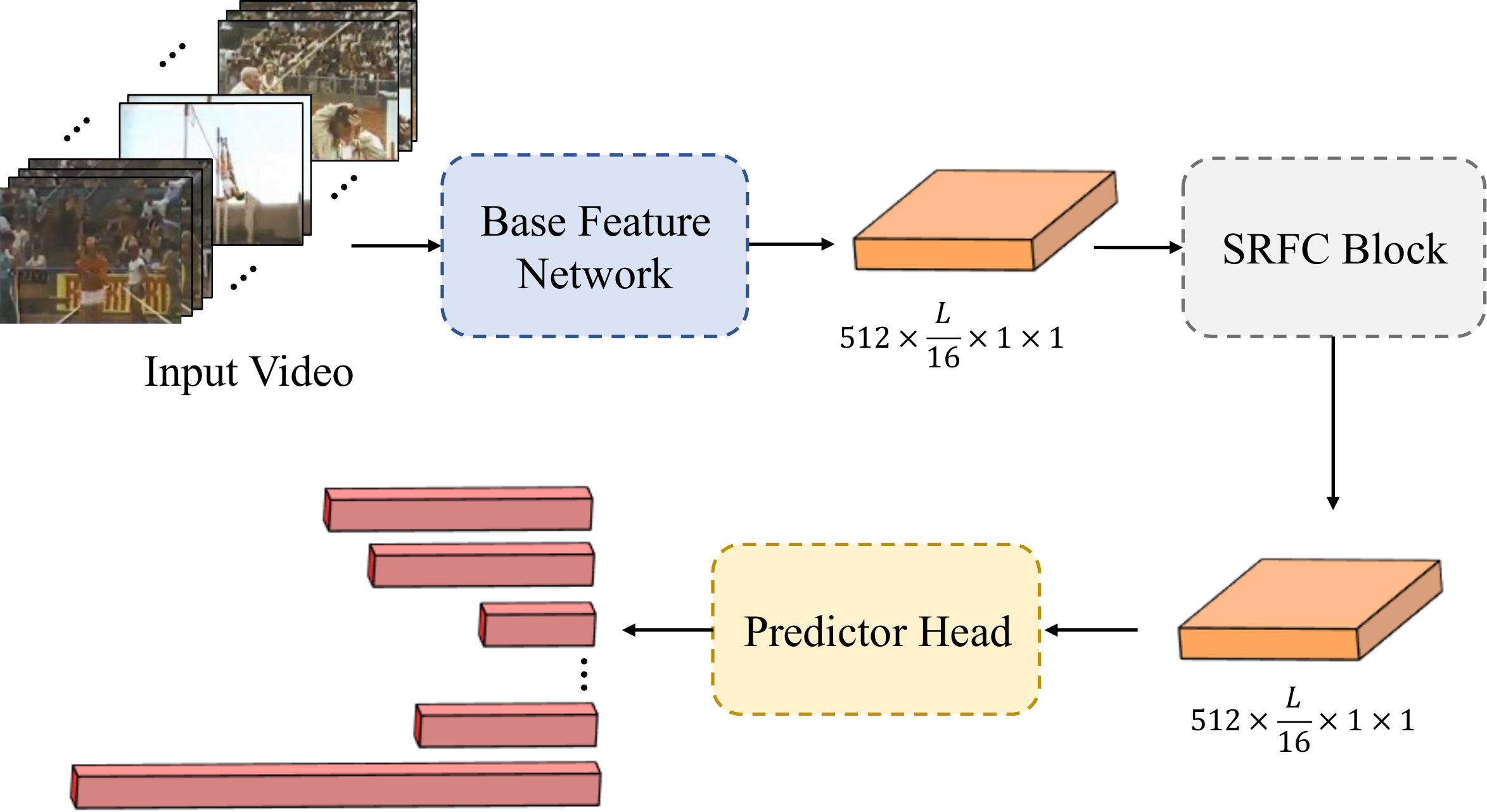}
    \vspace{-0.2cm}
    \caption{An overview of our SRF-Net architecture.}
    \label{fig:overview}
    \vspace{-0.2cm} 
\end{figure}
\section{Approach}
\label{sec:typestyle}
In this section, we present the proposed Selective Receptive Field Network (SRF-Net) in detail. Fig.\ref{fig:overview} illustrates an overview of our architecture. Firstly, the base feature network extracts a feature map from a sequential video clip. Then an SRFC block is utilized to dynamically adjust the receptive field size at each temporal location in the feature map. Finally, we exploit the predictor head for generating the classification scores and location offsets at each temporal location. The whole network is jointly optimized with action classification loss plus two regression losses, \ie, localization loss and center-ness loss, and can be trained in an end-to-end manner.
\subsection{Base Feature Network}
\label{ssec:bfn}
The input to our model is a sequence of RGB video frames with dimension $\mathbb{R}^{3\times L\times H\times W}$, where $3$ is the channel number of inputs, $L$ is the number of frames, $H$ and $W$ are the height and width of each frame. We adopt the convolutional layers ($conv1a$ to $conv5b$) of C3D \cite{tran2015learning} to generate a rich spatiotemporal feature map $\mathbf{C}_{conv5b}$ with dimension $\mathbb{R}^{512\times \frac{L}{8}\times \frac{H}{16}\times \frac{W}{16}}$. Then we feed the feature map into two 3D convolutional layers plus one max-pooling layer to extend its receptive field and produce a temporal feature map $\mathbf{T}$ with dimension $\mathbb{R}^{512\times \frac{L}{16}\times1\times1}$. For simplicity, the spatial dimension of $1\times1$ is omitted in all subsequent sections.
\subsection{Selective Receptive Field Convolution Block}
\label{ssec:srfc}
\begin{figure}[t]
    \centering
    \includegraphics[width=1 \linewidth]{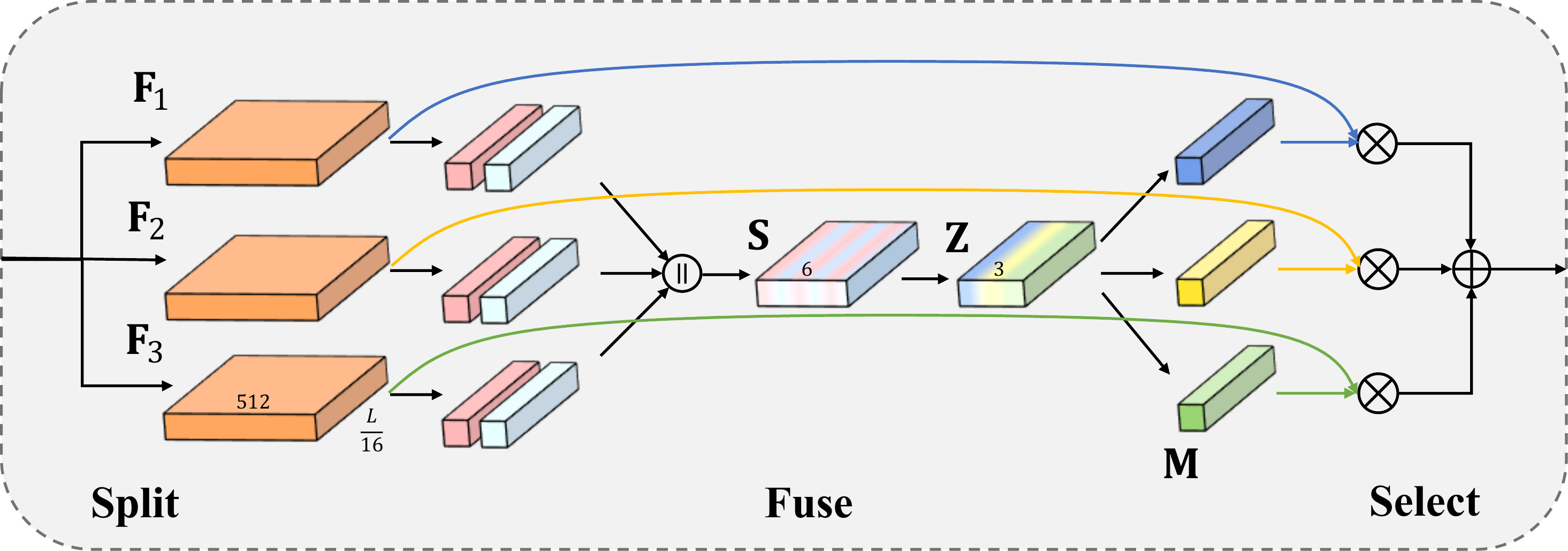}
    \vspace{-0.6cm}
    \caption{Selective Receptive Field Convolution (SRFC) Block.}
    \label{fig:SRFC Block}
    \vspace{-0.2cm} 
\end{figure}

In the SRFC block, multiple branches with different kernels are fused using softmax attention that is guided by the information in these branches. Different attentions on these branches yield different sizes of the effective receptive fields at locations in the temporal feature map. Inspired by dilated spatial convolution, we use dilated temporal convolution for multi-scale temporal feature learning. Specifically, as showed in Fig.\ref{fig:SRFC Block}, the SRFC block contains three operators—Split, Fuse and Select. 


\textbf{Split.} Given the temporal feature map $\mathbf{T}$, we first use three dilated temporal convolutional layers to produce three temporal feature maps $\mathbf{F}_1$, $\mathbf{F}_2$ and $\mathbf{F}_3$ with dilation rates increasing progressively to cover various temporal ranges. For $n$-th feature map $\mathbf{F}_n\in\mathbb{R}^{512\times \frac{L}{16}}$, the corresponding dilation rate is $r_n=2n-1$ to enlarge the temporal receptive fields efficiently. All dilated temporal convolutions have the same kernel size.
 
\textbf{Fuse.} Our base idea is to use an attention mechanism to control the information flows from multiple branches for each temporal location in the feature map. Therefore, we need to aggregate information from all branches firstly. 

Given the three temporal feature maps $\mathbf{F}_1$, $\mathbf{F}_2$ and $\mathbf{F}_3$, we apply average-pooling and max-pooling operations along the channel axis to embed channel information at each temporal location, generating two 1D maps: $\mathbf{S}_{n,avg}\in\mathbb{R}^{1\times\frac{L}{16}}$ and $\mathbf{S}_{n,max}\in\mathbb{R}^{1\times\frac{L}{16}}$ for each feature map $\mathbf{F}_n$ separately. Those are then concatenated along the channel axis to generate an efficient feature descriptor $\mathbf{S}\in\mathbb{R}^{6\times\frac{L}{16}}$:
\begin{equation}
    \setlength{\abovedisplayskip}{0.1cm}
\setlength{\belowdisplayskip}{0.1cm}
\begin{aligned}
\mathbf{S}_{n}&=[\mathbf{S}_{n,avg};\mathbf{S}_{n,max}]\\&=[AvgPool(\mathbf{F}_{n});MaxPool(\mathbf{F}_{n})], n=1,2,3.
\end{aligned}
    \label{eq:1}
\end{equation}

Next, on the concatenated feature descriptor $\mathbf{S}$, we apply temporal 1D convolutional layers to generate a soft attention map $\mathbf{M}\in\mathbb{R}^{3\times\frac{L}{16}}$, which enables the guidance for adaptive selection. Specifically, a softmax operator is applied along the channel axis:
\begin{equation}
    \setlength{\abovedisplayskip}{0.1cm}
\setlength{\belowdisplayskip}{0.1cm}
\mathbf{Z}=f^{k,3}\left(g\left(f^{k,d}\left(\mathbf{S}\right)\right)\right),\ \ \mathbf{Z}\in\mathbb{R}^{3\times\frac{L}{16}},
    \label{eq:2}
\end{equation}
\begin{equation}
    \setlength{\abovedisplayskip}{0.1cm}
\setlength{\belowdisplayskip}{0.1cm}
\mathbf{M}_n^x=\frac{e^{\mathbf{Z}_n^x}}{e^{\mathbf{Z}_1^x}+e^{\mathbf{Z}_2^x}+e^{\mathbf{Z}_3^x}},\ n=1,2,3,
    \label{eq:3}
\end{equation}
Where $f^{k,d}$ represents a convolution operation with the kernel size of $k$ and kernel number of $d$, $g$ is the ReLU activation function, $\mathbf{M}_1,\mathbf{M}_2,\mathbf{M}_3\in\mathbb{R}^{1\times\frac{L}{16}}$ are three soft attention vectors of $\mathbf{M}$, and $x$ indicates the temporal location in feature map. Note that $\mathbf{Z}_n^x$ is the $x$-th element of the $n$-th channel feature of $\mathbf{Z}$ and $\mathbf{M}_n^x$ is the $x$-th element of $\mathbf{M}_n$.

\textbf{Select.} The soft attention across channels is used to adaptively select the different temporal scale of information for each temporal location. Given three temporal feature maps $\mathbf{F}_1$,$\mathbf{F}_2$,$\mathbf{F}_3$ and the attention map $\mathbf{M}$, the final refined feature map $\mathbf{V}$ is computed as:
\begin{equation} 
    \setlength{\abovedisplayskip}{0.1cm}
\setlength{\belowdisplayskip}{0.1cm}
\mathbf{V}_i=\mathbf{M}_1\circ\mathbf{F}_{1,i}+\mathbf{M}_2\circ\mathbf{F}_{2,i}+\mathbf{M}_3\circ\mathbf{F}_{3,i}, 
    \label{eq:4}
\end{equation}
Where  $\mathbf{V}=\left[\mathbf{V}_{1\ };\mathbf{V}_{2\ };\ldots;\mathbf{V}_{512\ }\right],\ \ \mathbf{V}_i\in\mathbb{R}^{1\times\frac{L}{16}}$. 

\subsection{Predictor Head}
\label{ssec:ph}
Given the refined feature map $\mathbf{V}$ output from the SRFC block, two main branches are utilized to further classification and regression. The two main branches are realized by stacking two temporal 1D convolutional layers to predict action classification scores $\mathbf{c}$ and location offsets $\mathbf{t}$, respectively.

Different from anchor-based detectors, we directly regress the target temporal bounding box at the temporal location. Therefore, our detector directly views locations as training samples instead of anchor boxes in anchor-based detectors. Specifically, location $x$ in the temporal feature map is considered as a positive sample if it falls into any ground-truth action instance and the class label $c^\ast$ of the location is the class label of the ground-truth action instance. Otherwise, it is a negative sample and $c^\ast=0$ that indicates the background class. If a location falls into multiple action instances, we simply choose the action instance whose center point is most close to $x$ as its target. Therefore, besides the classification label, we get a label vector $\mathbf{t}^\ast=(l^\ast,r^\ast)$ being the regression targets for the location $x$, where the $l^\ast$ and $r^\ast$ are the duration from the location to the start and end time, respectively. 

\begin{figure}[t]
    \centering
    \includegraphics[width=0.8 \linewidth]{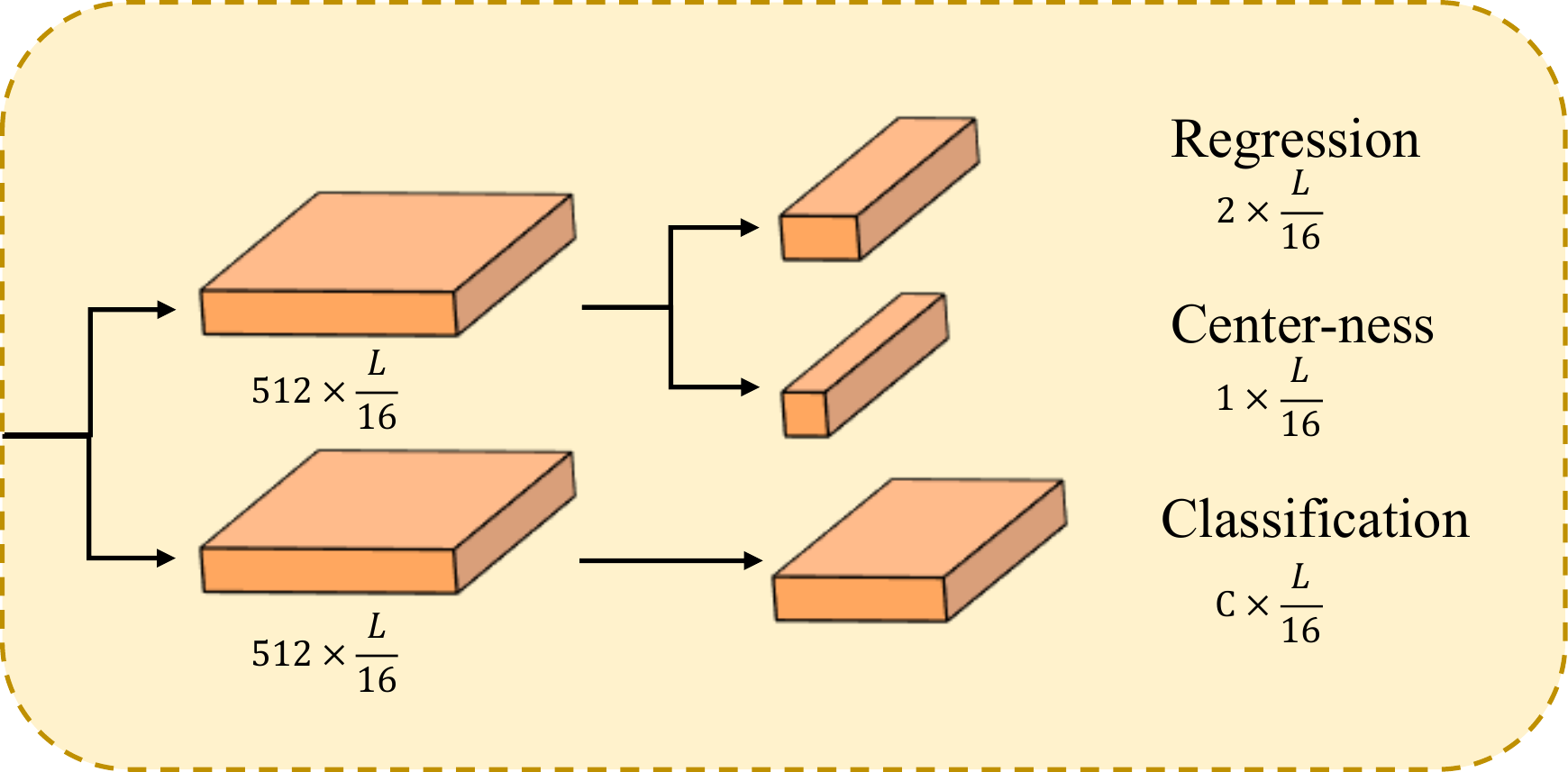}
    \vspace{-0.3cm}
    \caption{Predictor Head. $C$ is the number of action classes.}
    \label{fig:predictor head}
    \vspace{-0.2cm} 
\end{figure}

Inspired by FCOS \cite{tian2019fcos}, we use the similar effective strategy to suppress some low-quality action proposals produced by locations far away from the center of an action instance. Specifically, we add a single layer branch, in parallel with the regression branch as shown in Fig.\ref{fig:predictor head} to predict the “center-ness” of a location. The center-ness depicts the normalized distance from the location to the center of the action instance that the location is responsible for. Given the regression target $l^\ast$,$\ r^\ast$ for a location, the center-ness target is defined as:
\begin{equation}
    \setlength{\abovedisplayskip}{0.1cm}
\setlength{\belowdisplayskip}{0.1cm}
   {ctr}^\ast=\sqrt{\frac{\min(l^\ast,r^\ast)}{\max(l^\ast,r^\ast)}} .
    \label{eq:5}
\end{equation}
The center-ness ranges from 0 to 1. We adopt the center-ness score to down-weight the final prediction scores of proposals far away from the center of an action instance, which benefits the proposal re-ranking in the testing stage.
\subsection{Training and Inference}
\label{ssec:train&inference}

\textbf{Loss Function.} In the training stage, the overall training loss function in our SRF-Net is defined as a multi-task loss over all locations in the feature map $\mathbf{V}$ by integrating action classification loss, localization loss and center-ness loss:
\begin{equation}
    \setlength{\abovedisplayskip}{0.1cm}
\setlength{\belowdisplayskip}{0.1cm}
\begin{aligned}
Loss=&\frac{1}{N_{pos}}\sum_{x}{{Loss}_{cls}(\mathbf{c}_x,c_x^\ast})+\\&\frac{\lambda}{N_{pos}}\sum_{x}{\mathbb{I}_{{c_x^\ast>0}}{Loss}_{loc}(\mathbf{t}_x,\mathbf{t}_x^\ast})+\\&\frac{\beta}{N_{pos}}\sum_{x}{\mathbb{I}_{{c_x^\ast>0}}{Loss}_{ctr}({ctr}_x,{ctr}_x^\ast)}, 
\end{aligned}
    \label{eq:6}
\end{equation}
Where $\lambda$ and $\beta$ are the balance weight for localization loss and center-ness loss, and $N_{pos}$ denotes the number of positive samples. $\mathbb{I}_{{c_x^\ast>0}}$ is the indicator function, being 1 if $c_x^\ast>0$ and 0 otherwise. ${Loss}_{cls}$ is focal loss as in \cite{lin2017focal}. ${Loss}_{ctr}$ is a binary cross entropy (BCE) loss. ${Loss}_{loc}$ is the temporal IoU loss similar to UnitBox \cite{yu2016unitbox}. The temporal IoU loss is computed as:
\begin{equation}
    \setlength{\abovedisplayskip}{0.1cm}
\setlength{\belowdisplayskip}{0.1cm}
{Loss}_{loc}\left(\mathbf{t}_x,\mathbf{t}_x^\ast\right)=-\ln(\frac{{int}_x}{{union}_x}),
    \label{eq:7}
\end{equation}

where ${int}_x=\min{\left(l_x,l_x^\ast\right)}+\min{\left(r_x,r_x^\ast\right)}$ and ${union}_x=\left(l_x^\ast+r_x^\ast\right)+\left(l_x+r_x\right)-{int}_x$.

\textbf{Inference.} In the testing stage, the final ranking score of each action proposal is computed by multiplying the predict center-ness with the corresponding classification score. Finally, we filter out action proposals with final ranking scores below threshold $\alpha$ and further we apply the Non-Maximum Suppression (NMS) with the Jaccard overlap of $\delta$ to produce the detection results.

\begin{figure*}[t]
    \centering
    \includegraphics[width=1 \linewidth]{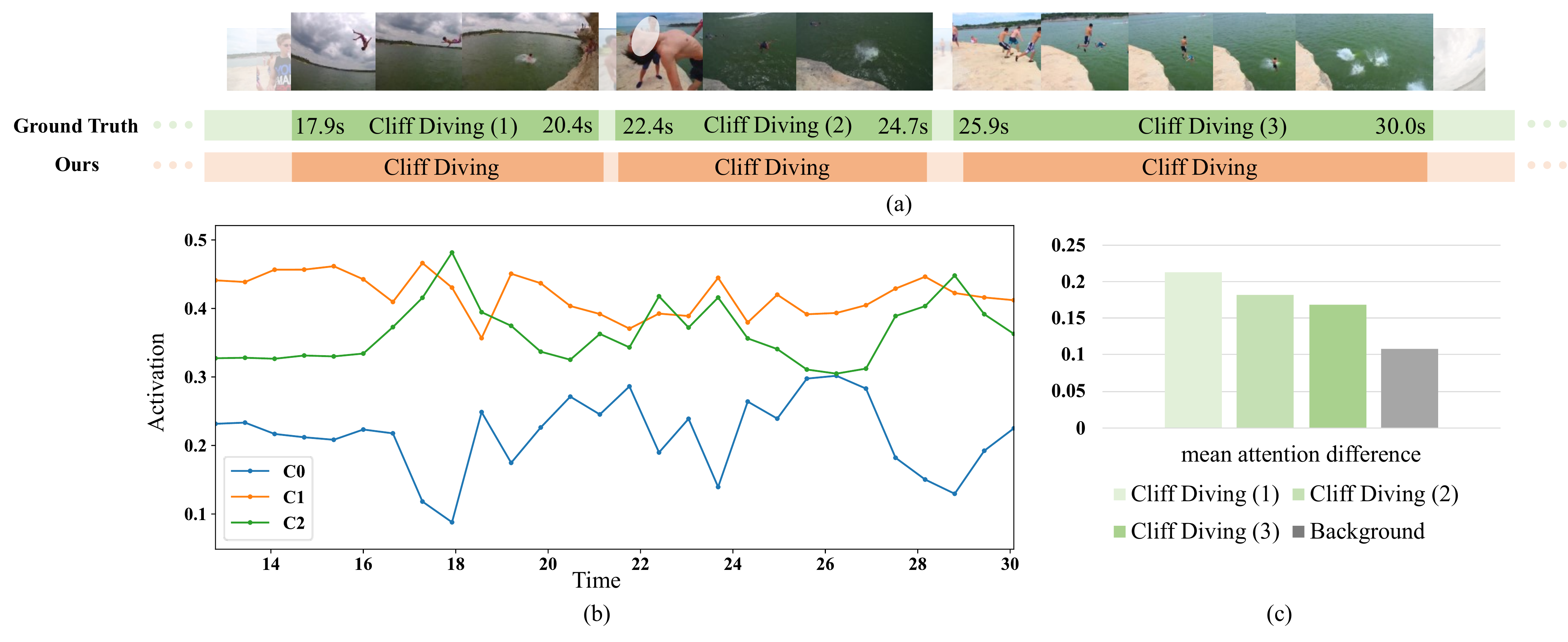}
    \vspace{-0.8cm}
    \caption{Qualitative results of temporal action detection by our SRF-Net on THUMOS14. Figure (a) visualizes predicted action instances on one example video clip from THUMOS14. The horizontal axis stands for time. Figure (b) shows attention values for multiple convolutional kernels in the corresponding period of this video clip predicted by our SRFC block. Figure (c) shows the mean attention value of C2 minus that of the C0 for three action instances.}
    \label{fig:visualize}
    \vspace{-0.1cm} 
\end{figure*}
\section{Experiments}
\label{sec:experiments}
\subsection{Implementation Details}
\label{ssec:expsetting}
\textbf{Dataset.} We conduct experiments on the THUMOS14 dataset. THUMOS14 contains 1,010 untrimmed videos for validation and 1,574 untrimmed videos for testing from 20 sports action classes. Among all the videos, there are 220 and 212 videos with temporal annotations in the validation and testing set, respectively. The dataset is particularly challenging as it consists of very long videos with multiple action instances of diverse duration. This is in contrast to many other datasets such as ActivityNet \cite{caba2015activitynet}, which only has on average $\sim$1 action instance per video. Following \cite{zhao2017temporal}, the validation set is used for training and the testing set is used for evaluation. We report the mean Average Precision (mAP) metric (\%) with IoU thresholds varied from 0.3 to 0.7. 

\textbf{Network Details.} SRF-Net takes as input $L=768$ raw video frames with size $H=W=112$. We decode each video at 25 FPS. Then we generate a series of video clips by sliding window. We use C3D weights pre-trained on UCF101 to initialize the base feature network. The initial learning rate is 0.0006 and decreased by 10\% after every 4 epochs. The batch size is 2. The balance wight $\lambda$ and $\beta$ are all set as 1. In inference, the threshold $\alpha$ is 0.005 and the NMS threshold $\delta$ is 0.3 and the top 300 confident predictions are picked as the final detection results.
\subsection{Ablation Study}
\label{ssec:ablation}
To investigate the effect of designed components and settings in our SRF-Net, we conduct ablation studies on THUMOS14.

\textbf{Effect of SRFC Block.} We investigate the effectiveness of the SRFC block and how each design in the SRFC block influences the overall performance. 
We use three different convolutional kernels in the split operation. “C0” is standard convolution with kernel size 3. “C1” and “C2” is the dilated convolution with dilation size 3; and dilation size 5, respectively. Table \ref{table:ablation1} details the mAP performance by considering different combinations of multiple branches. The “SRF” means that we use the attention mechanism, otherwise we simply sum up the results with these dilated convolution kernels. 
It can be seen that the multi-branch manner outperforms each single branch manner and the detection performance is boosted as the attention mechanism is used (the mAP@0.5 improves from 43.2\% to 44.8\%). 
The results validate the effectiveness of the proposed SRFC block and demonstrate that multi-scale temporal information and adaptive adjustment of the receptive field is important for tackling actions with different scales.

\begin{table}[t]
    \caption{Result of SRF-Net with different combinations of multiple branches.}
    \vspace{0.1cm}
	\label{table:ablation1}
	\centering

  \begin{tabular}{c|ccc|cc}
\toprule
\multicolumn{1}{l|}{} & \multicolumn{3}{c|}{single branch} & \multicolumn{2}{l}{multi-branch}  \\ 
\multicolumn{1}{l|}{} & (a)  & (b)  & (c)                  & (d)  & (e)\\ 
\midrule
C0                    & \checkmark     &      &                      & \checkmark     & \checkmark\\
C1                    &      & \checkmark     &                      & \checkmark      & \checkmark\\
C2                    &      &      & \checkmark                     & \checkmark     & \checkmark\\
SRF                   &      &      &                      &      & \checkmark\\ 
\midrule
mAP@0.5               & 38.9 & 38.9 & 36.3                 & 43.2 & \textbf{44.8}\\ 
mAP@0.7               & 17.1 & 18.1 & 16.4                 & 19.2 & \textbf{20.9 }\\
\bottomrule

\end{tabular}
\vspace{-0.1cm} 
\end{table}

\begin{table}[t]
        \caption{Ablation study for the center-ness branch.}
        \vspace{0.1cm}
        \label{table:ablation2}
        \centering
    
      \begin{tabular}{c|cc}
    \toprule
    Method                     & mAP@0.5       & mAP@0.7        \\ 
    \midrule
    None                       & 41.5          & 16.8           \\
    center-ness w/o prediction & 41.7          & 16.8           \\
    center-ness                & \textbf{44.8} & \textbf{20.9}  \\
    \bottomrule
    
    \end{tabular}
    \vspace{-0.1cm} 
    \end{table}

\textbf{Effect of Center-ness.} The results in the fourth row of Table \ref{table:ablation2} show that the center-ness branch is very helpful to boost the detection performance (the mAP@0.5 improves from 41.5\% to 44.8\%). Note that center-ness can also be computed from the predicted regression vector without the extra center-ness branch. However, the results in the third row 
of Table \ref{table:ablation2} 
show that this manner cannot really improve the performance and thus the separate center-ness branch is necessary.

\subsection{Qualitative Results}
\label{ssec:QR}
We provide qualitative results to demonstrate the effectiveness of our proposed SRF-Net. As shown in Fig.\ref{fig:visualize}, our SRF-Net accurately detects three action instances in the video clip and the attention values are adaptively adjusted as actions occur. It is seen that at the time of the Cliff Diving action, the difference between the mean attention weights associated with C0 (smaller) and C2 (larger) is greater than that of the background, which suggests that as these actions occur, more attention is assigned to larger kernels.
\subsection{Comparisons with State-of-the-Art}
\label{ssec:sota}
We compare with several state-of-the-art techniques of TAD on THUMOS14. Table \ref{table:sota} lists the mAP performances with different IoU thresholds. In comparison, our SRF-Net shows superior performance than other anchor-based approaches, such as R-C3D \cite{xu2017r} and SSAD \cite{lin2017single}, and our SRF-Net does not need to pre-design anchor on input videos with different action durations. Using simple C3D network as backbone, our SRF-Net exhibits better performance than other one-stage approaches and is comparable with the state-of-the-art two-stage approaches. In particular, our SRF-Net achieves 44.8\% of mAP@0.5, which outperforms the one-stage competitor GTAN \cite{long2019gaussian} by 6.0\% and the best two-stage approach G-TAD \cite{xu2020g} by 4.6\%. 
Such results demonstrate the superiority of our SRF-Net.

\begin{table}[t]
    \caption{Temporal action detection results on THUMOS14.}
    \vspace{0.1cm}
	\label{table:sota}
	\centering
  \setlength{\tabcolsep}{1.5mm}
\begin{tabular}{c|ccccc} 
\toprule
Approach           & 0.3           & 0.4           & 0.5           & 0.6           & 0.7            \\ 
\midrule
\multicolumn{6}{c}{Two-stage Temporal Action Detection}                                                      \\ 
\midrule
CDC (2017)\cite{shou2017cdc}     & 40.1          & 29.4          & 23.3          & 13.1          & 7.9            \\
TURN (2017)\cite{gao2017turn}     & 44.1          & 34.9          & 25.6          & 14.6          & 7.7            \\
R-C3D (2017)\cite{xu2017r}    & 44.8          & 35.6          & 28.9          & -             & -              \\
SSN (2017)\cite{zhao2017temporal}     & 51.9          & 41.0          & 29.8          & -             & -              \\
BSN (2018)\cite{lin2018bsn}      & 53.5          & 45.0          & 36.9          & 28.4          & 20.0           \\
BMN (2019)\cite{lin2019bmn}     & 56.0          & 47.5          & 38.8          & 29.7          & 20.5           \\
DBG (2020)\cite{lin2020fast}     & \textbf{57.8} & 49.4 & 39.8          & 30.2          & 21.7           \\
FC-AGCN-P-C3D (2020) \cite{li2020graph} & 57.1 & \textbf{51.6}  & 38.6          & 28.9          & 17.0 \\
G-TAD (2020)\cite{xu2020g}   & 54.5          & 47.6          & \textbf{40.2} & \textbf{30.8} & \textbf{23.4}  \\ 

\midrule
\multicolumn{6}{c}{One-stage Temporal Action Detection}                                                      \\ 
\midrule
Yeung et al. (2016)\cite{yeung2016end} & 36.0 &26.4 &17.1 &- &- \\
SSAD (2017)\cite{lin2017single}     & 43.0          & 35.0          & 24.6          & -             & -              \\
SS-TAD (2017)\cite{DBLP:conf/bmvc/BuchEGFN17}   & 45.7          & -             & 29.2          & -             & 9.6             \\
AFO-TAD (2019)\cite{tang2019afo} & 56.4          & 50.6          & 42.0          & 31.2          & 19.6           \\
GTAN (2019)\cite{long2019gaussian}     & \textbf{57.8} & 47.2          & 38.8          & -             & -              \\ 
\midrule
SRF-Net (Ours)     & 56.5          & \textbf{50.7} & \textbf{44.8} & \textbf{33.0} & \textbf{20.9}  \\
\bottomrule
\end{tabular}
\vspace{-0.1cm} 
\end{table}
\section{Conclusions}
\label{sec:conclusion}
In this paper, we proposed a novel Selective Receptive Field Network (SRF-Net) for TAD task. The SRF-Net directly predicts location offsets and classification scores at each temporal location in the feature map without pre-defined anchors, which benefits from a new designed building block called Selective Receptive Field Convolution (SRFC). The SRFC block can adaptively adjust its receptive field size at each temporal location in a soft-attention manner. Experimental results demonstrate the effectiveness and contribution of our SRFC block, which brings \textbf{substantial} improvements and our SRF-Net model achieves superior performance than state-of-the-art TAD approaches on THUMOS14.


\bibliographystyle{IEEEbib}

\end{document}